\documentclass[11pt,a4paper]{article}
\usepackage{times,latexsym}
\usepackage{url}
\usepackage[T1]{fontenc}

\usepackage{graphicx}

\usepackage{xcolor}         % colors
\usepackage{amsmath}
\usepackage{multirow}
\usepackage{booktabs}
\usepackage{diagbox} 
\usepackage{subcaption}
\usepackage[normalem]{ulem} 
\usepackage{footmisc}

\usepackage[acceptedWithA]{tacl2021v1}
\usepackage{xspace,mfirstuc,tabulary}

\newif\iftaclinstructions
\taclinstructionsfalse % AUTHORS: do NOT set this to true
\iftaclinstructions
\renewcommand{\confidential}{}
\renewcommand{\anonsubtext}{(No author info supplied here, for consistency with
TACL-submission anonymization requirements)}
\newcommand{\instr}
\fi

\iftaclpubformat % this "if" is set by the choice of options

\else

\fi

\title{What Defines Good Reasoning in LLMs?\\Dissecting Reasoning Steps with Multi-Aspect Evaluation}

\author{
Heejin Do\textsuperscript{1,2}\Thanks{Work done during an internship at NAVER AI Lab.} \quad
Jaehui Hwang\textsuperscript{2} \quad
Dongyoon Han\textsuperscript{2} \quad
Seong Joon Oh\textsuperscript{3} \quad
Sangdoo Yun\textsuperscript{2}\Thanks{Corresponding author.}\\
\textsuperscript{1}ETH Zürich, ETH AI Center \quad
\textsuperscript{2}NAVER AI Lab \quad
\textsuperscript{3}University of Tübingen\\
\texttt{heejin.do@ai.ethz.ch}
}

\date{}

\begin{document}
\maketitle

\begin{abstract}
Evaluating large language models (LLMs) on final-answer correctness is the dominant paradigm. This approach, however, provides a coarse signal for model improvement and overlooks the quality of the underlying reasoning process. We argue that a more granular evaluation of reasoning offers a more effective path to building robust models. We decompose reasoning quality into two dimensions: \textit{relevance} and \textit{coherence}. Relevance measures if a step is grounded in the problem; coherence measures if it follows logically from prior steps. To measure these aspects reliably, we introduce causal stepwise evaluation (CaSE). This method assesses each reasoning step using only its preceding context, which avoids hindsight bias. We validate CaSE against human judgments on our new expert-annotated benchmarks, MRa-GSM8K and MRa-MATH. More importantly, we show that curating training data with CaSE-evaluated relevance and coherence directly improves final task performance. Our work provides a scalable framework for analyzing, debugging, and improving LLM reasoning, demonstrating the practical value of moving beyond validity checks.
\end{abstract}
\section{Introduction}

Reasoning is a critical capability for large language models (LLMs) \cite{wei2022chain,welleck2022naturalprover,hao2024llm,zhang2024careful,li2024common}. Recent advances in LLM reasoning have been achieved with reinforcement learning~\cite{cui2025process,xiong2025self,ren2025deepseek} and search-based strategies~\cite{luo2024improve,li2025enhancing,lin2025cmcts,ma2025step}, both of which fundamentally rely on a precise evaluation of reasoning to provide reward signals and guide the search.

Yet evaluation of reasoning capability has predominantly focused on final-answer correctness, overlooking the quality of the reasoning process. While simple and useful for tracking high-level progress, that metric is too coarse as a signal for improvement: it certifies outcomes but reveals little about the process that produced them. Recent work on step-level supervision \cite{lightman2023lets,song2025prmbench} and meta-reasoning benchmarks \cite{zeng2023mr,zeng2024mr,xia2025evaluating} moves beyond outcomes, but largely defines reasoning quality as step correctness \cite{lee2025evaluating}. However, even locally valid steps can be irrelevant to the goal or incoherent as a chain, making correctness an insufficient criterion; optimizing for it alone risks redundant steps and non-causal progressions.

\begin{figure}
    \centering
    \includegraphics[width=\linewidth]{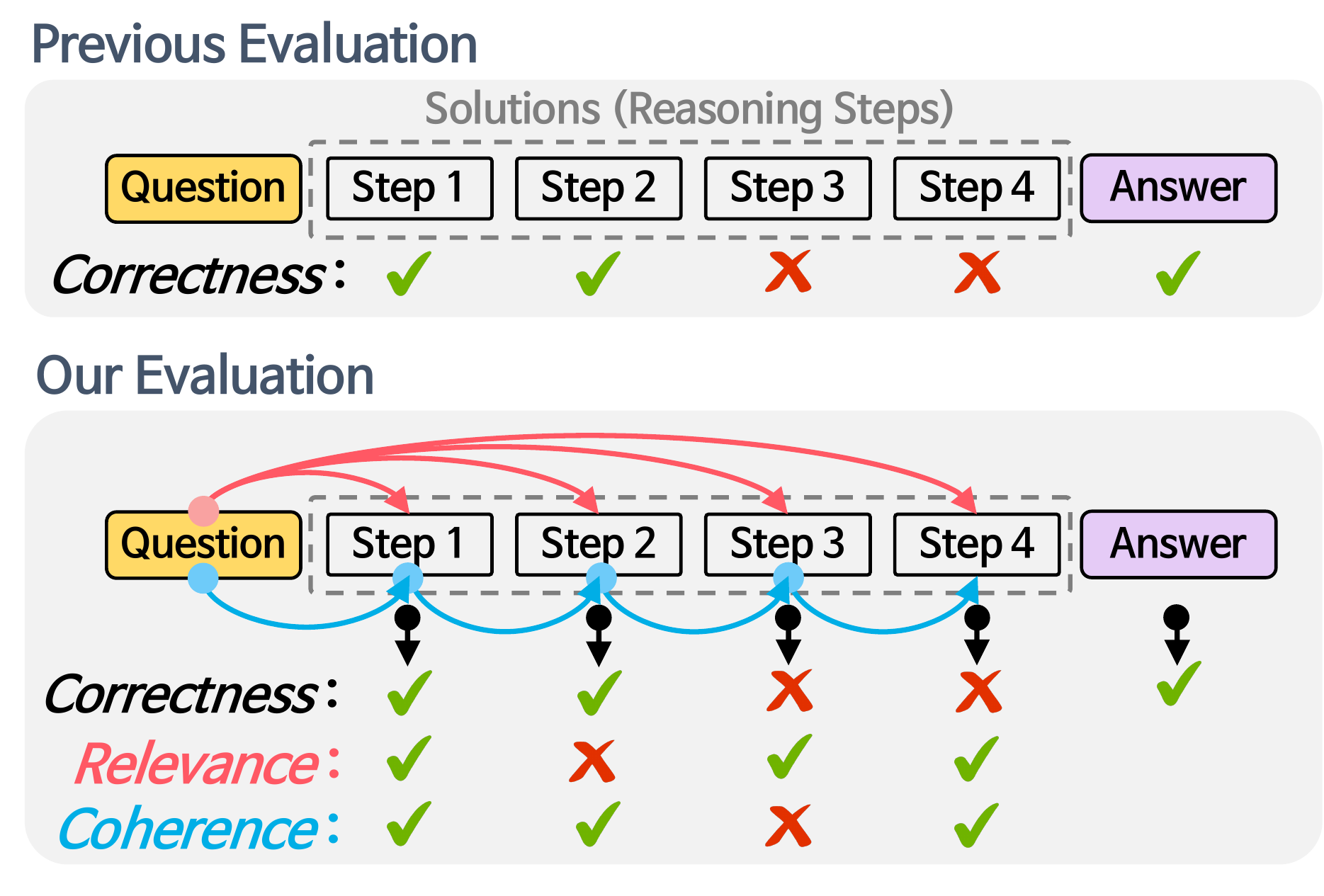}
    \caption{Previous evaluations emphasize step- or answer-level \textit{correctness}~\cite{lightman2023lets,song2025prmbench,zeng2023mr,zeng2024mr}. We formalize two causal dimensions at the step-level: \textcolor[RGB]{255, 90, 103}{\textit{relevance}} (grounding to the question) and \textcolor[RGB]{10, 175, 229}{\textit{coherence}} (logical consistency with prior steps).
    }
    \label{fig:fig1}
\end{figure}

In this work, we formalize reasoning quality with two dimensions beyond correctness: relevance and coherence (Figure~\ref{fig:fig1}). Relevance assesses whether a step is well-grounded in and addresses the problem, and coherence assesses whether a step logically follows from the preceding steps. Just as human learners deepen understanding by reflecting on their reasoning beyond correctness \cite{herbert2022engagement,mwamba2019effects,smit2017effects}, we posit that LLMs similarly benefit from more granular, process-level evaluation.

To enable validation of diverse LLMs' ability to judge \textit{relevance} and \textit{coherence}, we construct multi-aspect, step-level meta-reasoning benchmarks: {MRa-GSM8K} and {MRa-MATH}. The LLM-generated solution traces are segmented into intermediate steps and annotated by human experts for \textit{relevance} and \textit{coherence}, complementing the step correctness labels provided in the underlying meta-reasoning datasets \cite{zeng2023mr, xia2025evaluating}. Analyses on our benchmarks reveal that, even among traces with incorrect steps, those that maintain both \textit{relevance} and \textit{coherence} are more likely to reach the correct final answer. These findings position the proposed aspects as complementary, process-level evidence of \textit{good reasoning}, underscoring the importance of measuring them.

The remaining question is how to evaluate these aspects at scale. However, most evaluation methods compared on meta-reasoning benchmarks are LLM-as-a-judge protocols \cite{zeng2024mr, zeng2023mr}, which typically score entire traces at once or condition on future steps. These correctness-centric approaches potentially induce hindsight with causality and make them ill-suited for measuring \textit{relevance} and \textit{coherence}. Therefore, we introduce Causal Stepwise Evaluation (CaSE), an evaluation strategy that emulates the causal auto-regressive generation process by scoring each step solely based on its preceding context. Across multiple LLM judge backbones and datasets, CaSE achieves significantly stronger agreement with human annotations than whole-trace baselines.

Beyond measurement, supporting step-level evaluation of \textit{relevance} and \textit{coherence} yields practical advantages, which we demonstrate with {CaSE}. First, in supervised fine-tuning (SFT) data curation, CaSE scores provide an explicit criterion to filter out low-quality steps, yielding measurable gains in downstream accuracy, outperforming heuristic filtering methods such as s1~\citep{muennighoff2025s1}. 
Second, our analysis of reasoning traces shows that guiding the generation process with CaSE (\textit{e.g.}, SFT data curation, or CaSE-guided inference) leads to higher-quality reasoning.
Taken together, these results position our multi-aspect evaluation framework as a unified approach to \textit{evaluate}, \textit{analyze}, and \textit{improve} LLM reasoning. Our contributions are:

\begin{itemize}
\item Establish relevance and coherence as key dimensions for step-level reasoning evaluation

\item Release MRa-GSM8K and MRa-MATH with human expert step-level annotations for relevance and coherence, and analyses linking these aspects to problem-solving success.

\item Introduce CaSE, a causal stepwise evaluation that avoids hindsight bias and achieves higher agreement with human judgments.

\item Demonstrate practical utility of multi-aspect evaluation via CaSE for SFT curation and inference-time guidance.

\end{itemize}
\section{Related Work}
\paragraph{Meta-reasoning benchmarks}

\begin{figure*}
    \centering
    \includegraphics[width=0.98\linewidth]{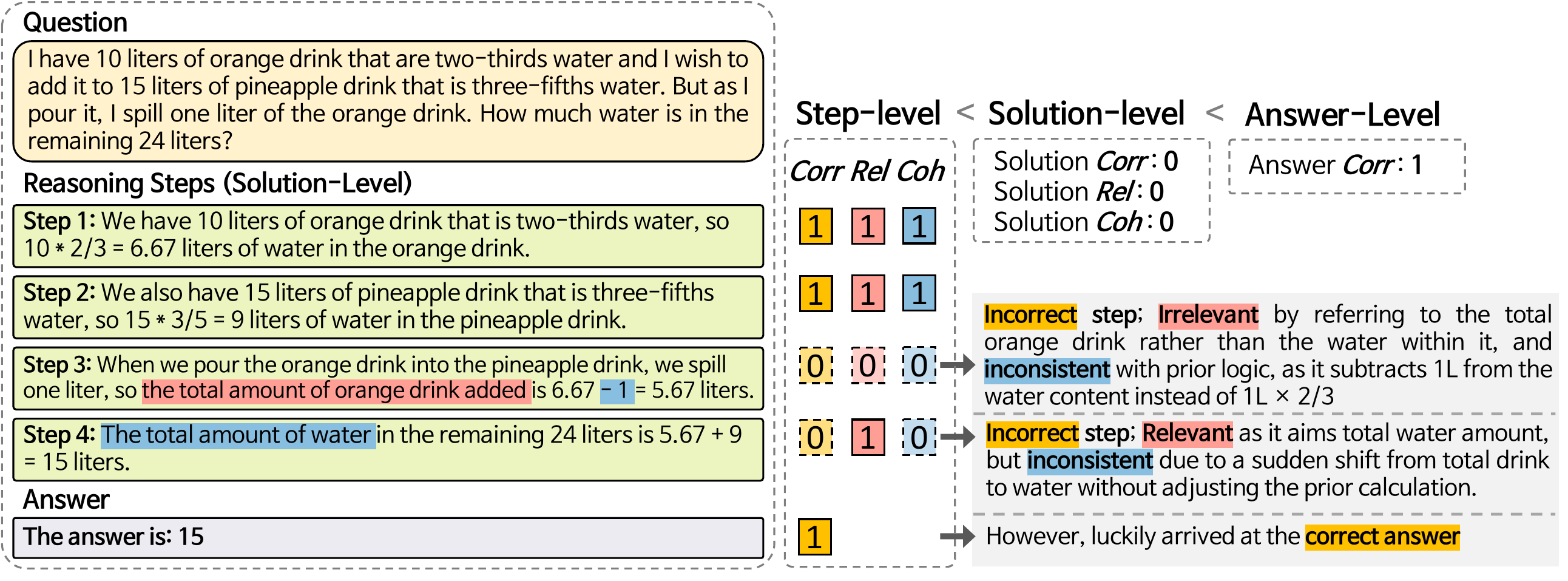}
    \caption{An example of the proposed fine-grained multi-aspect evaluation. We enable more granular diagnosis of reasoning quality by shifting from answer-level to step-level evaluation and extending the criteria beyond correctness to include relevance and coherence. \textit{Corr}, \textit{Rel}, and \textit{Coh} denote \textit{Correctness}, \textit{Relevance}, and \textit{Coherence}, respectively. Dotted boxes indicate the real datasets; the gray boxes on the right are explanations we added to aid understanding.}
    \label{fig:overview}
\end{figure*}

Although LLM reasoning abilities are often judged by the final-answer accuracy~\citep{cobbe2021gsm8k,lightman2023lets,rein2024gpqa}, recent work emphasizes that correct answers do not necessarily imply valid reasoning traces~\citep{wei2022chain, zelikman2022star}. This limitation has motivated a shift from outcome-based evaluation toward reasoning process evaluation, resulting in the emergence of \textit{meta-reasoning} benchmarks. Benchmarks such as GSM-Symbolic~\citep{mirzadeh2024gsm}, MR-Ben~\citep{zeng2024mr}, and PROCESSBENCH~\citep{zheng2024processbench} evaluate the model’s ability to reflect on, verify, or assess reasoning chains. \citet{zeng2023mr} and \citet{xia2025evaluating} propose MR-GSM8K and MR-MATH, respectively, by collecting LLM-generated reasoning steps for solving GSM8K~\cite{cobbe2021gsm8k} and MATH500~\citep{lightman2023lets} problems and human-labeling entire solution-level accuracy to address the existing model's reliance on final-answer-only evaluation. However, even robust LLMs struggle to detect flawed reasoning steps, highlighting the need for specified methods for step-level evaluation. Further, they emphasize correctness (and also redundancy for efficiency in \citet{xia2025evaluating}) but overlook contextual fit and causal consistency. We address this gap by formalizing two dimensions, \textit{relevance} and \textit{coherence}.

\paragraph{Reasoning evaluation beyond correctness}
To move beyond correctness, ROSCOE~\citep{golovneva2022roscoe} introduces multiple evaluation dimensions such as grammar, factuality, redundancy, and coherence. However, it relies on reference-based comparisons to ground-truth reasoning chains, limiting its ability to incorporate the diversity of valid reasoning paths. \citet{jacovi2024chain} benchmarks verifiers in open-domain QA with evidence-grounded, answer-centric labels (relevance to the final answer and logical correctness) and finds that current LMs struggle to judge them. THINK-Bench~\citep{li2025think} and MME-CoT~\citep{jiang2025mme} extend evaluation to dimensions such as efficiency and robustness to measure overthinking behavior of LLMs; however, the per-step semantic quality measurement remains coarse. Some recent studies attempt to diversify evaluation for process reward models (PRMs), e.g., \citet{song2025prmbench} categorizes PRMs' error taxonomies into simplicity, soundness, and sensitivity. However, they primarily evaluate PRMs' capacity rather than assessing the LLM-generated reasoning traces themselves. 
In contrast, we introduce a reference-free framework that directly measures the semantic quality and causal flow of individual steps, providing a multi-view of how traces drive final outcomes.

\enlargethispage{\baselineskip}
%\section{Revisiting Reasoning Quality}

\section{Multi-aspect Reasoning Evaluation}

\subsection{Dissecting Reasoning Quality}

\paragraph{Pedagogical insights}

Progress in human learning rarely stems from knowing whether an answer is right; instead, it emerges from understanding how a solution unfolds and why it works or fails. This principle has long shaped performance assessment in mathematics education, where evaluators prioritize reasoning processes over final answers \cite{herbert2022engagement, mwamba2019effects, smit2017effects}. Across instructional contexts, especially in U.S. school systems, structured rubrics often assess three core dimensions: problem interpretation, solution planning, and justification of the final outcome \cite{loong2018developing, shirawia2024performance}. These multi-dimensional evaluations provide fine-grained feedback that promotes students’ deeper understanding. We argue that reasoning in LLMs deserves a similar lens, evaluated not as a single outcome, but as a process with interpretable structure within a pedagogically grounded evaluation framework.

\paragraph{Evaluation aspects}

Informed by established practices in mathematics education \cite{loong2018developing, shirawia2024performance}, where student reasoning is evaluated not merely by correctness but by its alignment with the problem context and the logical structure of the solution path, we adopt two foundational criteria for step-level reasoning evaluation in LLMs: \textit{relevance} and \textit{coherence}.

\begin{itemize}
\item \textbf{Relevance} assesses whether a step is well-grounded in the question and addresses a necessary part of the solution, i.e., meaningfully contributes to solving the given problem.

\item \textbf{Coherence} reflects whether a step logically follows from the preceding steps, forming a consistent chain of reasoning.

\end{itemize}

\begin{figure}[t]
    \centering
    \includegraphics[width=\linewidth]{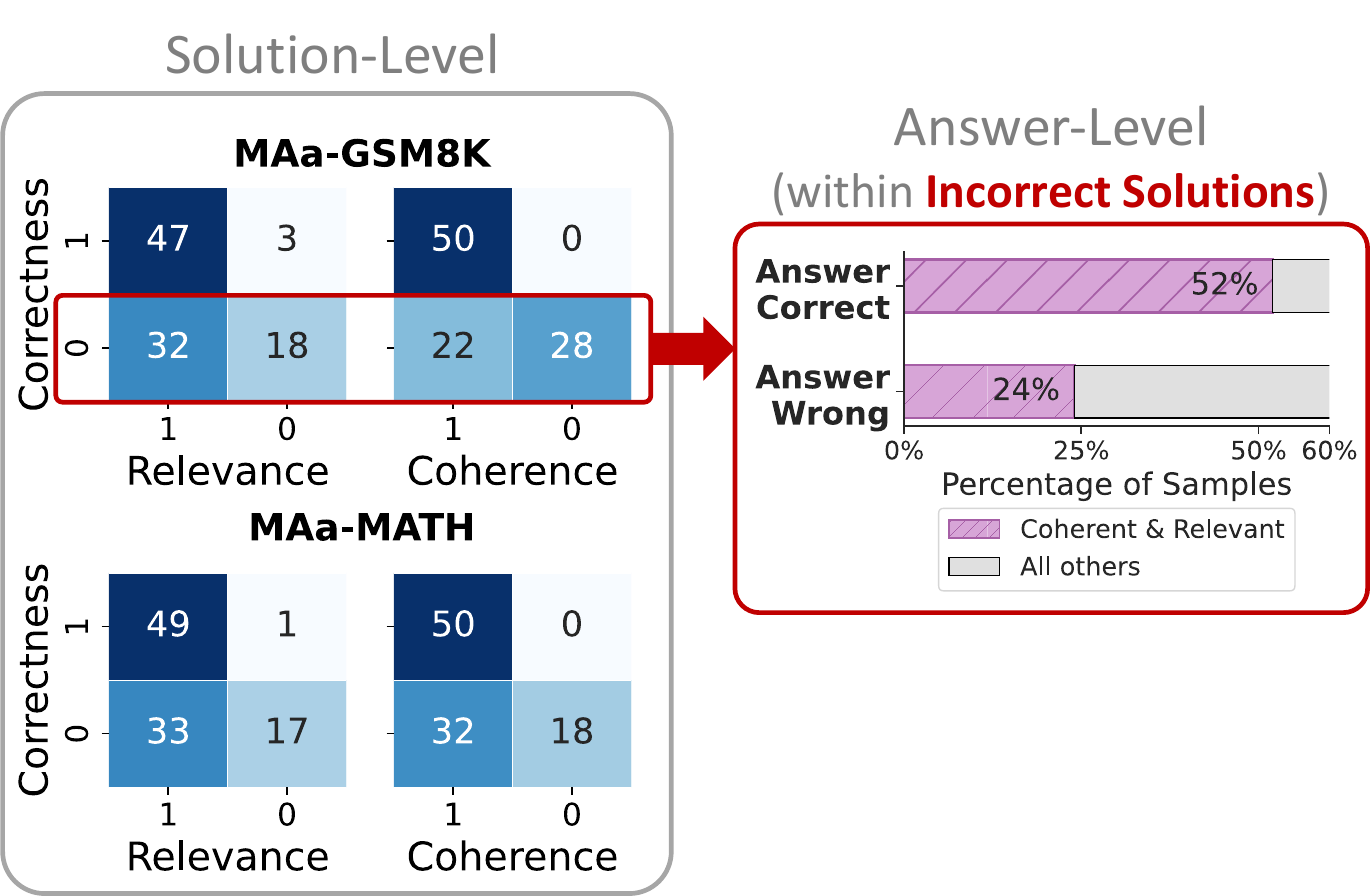}
    \caption{Confusion between \textbf{solution-level} correctness and relevance/coherence labels (left), and breakdown of incorrect solutions in MRa-GSM8K by \textbf{answer outcome}, highlighting the proportion satisfying both relevance and coherence (right).
    }
    \label{fig:bench1}
\end{figure}

\subsection{Constructing Multi-Aspect Benchmarks}
To support validation of diverse models' ability to judge \textit{relevance} and \textit{coherence}, we construct two multi-aspect, step-level meta-reasoning benchmarks: \textbf{MRa-GSM8K} and \textbf{MRa-MATH}, which extend prior meta-reasoning datasets MR-GSM8K~\cite{zeng2023mr} and MR-MATH~\cite{xia2025evaluating}.
The prior datasets provide human judgments of solution-level correctness (i.e., trace-level validity) and final-answer correctness for LLM-generated solutions to GSM8K and MATH. For solution generation, math-oriented models such as MetaMath~\cite{yu2023metamath}, Abel~\cite{abel}, and WizardMath~\cite{luo2023wizardmath} were used. We randomly sample 100 question-solution pairs from each dataset with a 1:1 ratio of solution-level correct vs.\ incorrect.

\paragraph{Annotation process}
We recruited six mathematics education experts through the Upwork\footnote{\url{https://www.upwork.com/}} platform, who have expertise in teaching mathematics. Each annotator was assigned 100 problems and performed binary labeling of \textit{relevance} and \textit{coherence} on every reasoning step, following detailed guidelines. For each benchmark, three independent annotators labeled all samples to ensure labeling consistency. We refer to the assessment conducted at the step level as \textbf{step-level} evaluation. For either aspect (relevance or coherence), the \textbf{solution-level} score is 1 only if all steps satisfy the aspect; if any step fails, the score is 0. The \textbf{answer-level} score reflects only the correctness of the final answer (Figure~\ref{fig:overview}). After completing annotations, we assessed the perceived utility by asking annotators whether evaluating relevance and coherence separately from the correctness offered meaningful insight into reasoning quality. Five out of six respondents affirmed its importance, with one neutral, supporting the value of our proposed evaluation dimensions.

\section{Empirical Evidence: Relevance and Coherence Drive Successful Reasoning}

\subsection{Observations from Datasets}\label{sec4.1}

\begin{figure}[t]
    \centering
    \includegraphics[width=\linewidth]{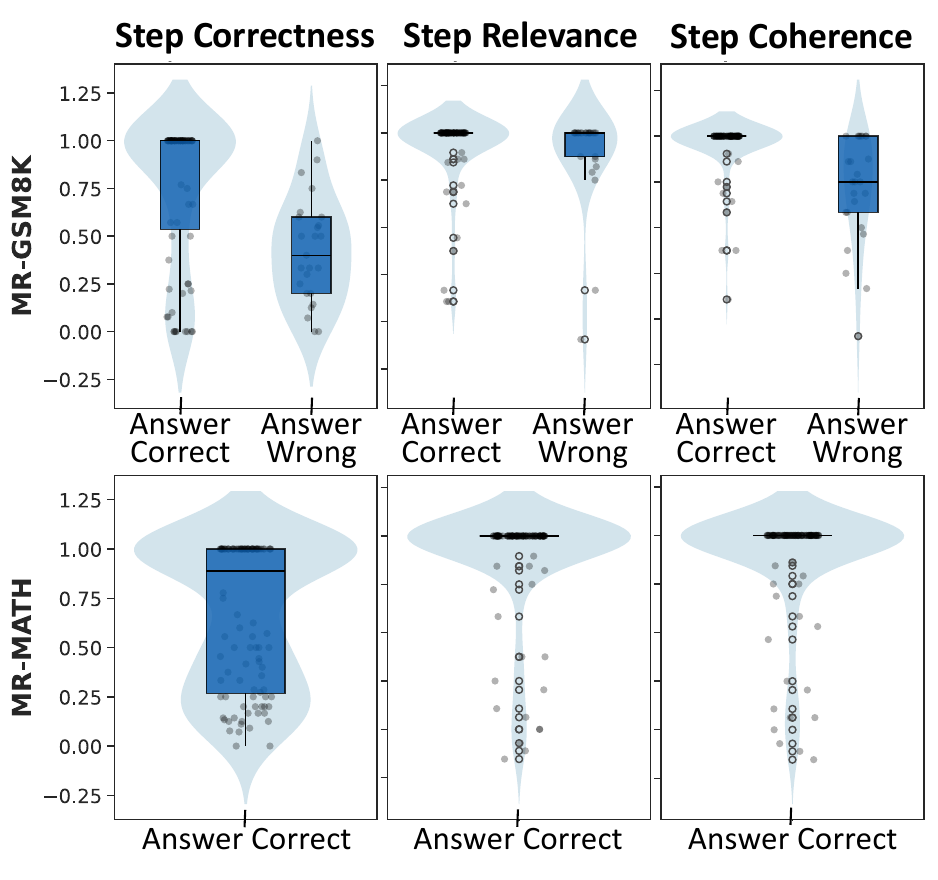}
    \caption{The distribution of average step-level \textit{Correctness}, \textit{Relevance}, and \textit{Coherence} across solutions (y-axis), grouped by answer-level correctness (x-axis).
    }
    \label{fig:bench2}
\end{figure}

We examine the relationship between the solution-level correctness labels from the original meta-reasoning datasets and our newly annotated relevance and coherence dimensions. As shown in Figure~\ref{fig:bench1} (left), correct solutions mostly satisfy both criteria, while a substantial portion of incorrect solutions also exhibit one or both. To better understand such cases, we take a closer look at MRa-GSM8K, restricting to solutions that are solution-level incorrect (i.e., contain at least one invalid step). We find that incorrect solutions whose steps are nevertheless solution-level relevant and solution-level coherent (all steps satisfy relevance and coherence) are more than twice as likely to yield the correct final answer as those that violate either aspect (52\% vs. 24\%).

These findings highlight that the defined aspects do more than assess internal reasoning quality; they also serve as potential signals of problem-solving success. %success trajectories. 
Note that the original MR-MATH dataset contains only samples with correct final answers, though their solutions may be correct or incorrect; thus, the right-hand analysis is conducted solely on the MRa-GSM8K.

To obtain more precise proportions, we examined the average of step-level labels instead of assigning a solution-level label of 0 whenever any step is labeled 0. 
Figure~\ref{fig:bench2} shows that when the final answer is correct, step-level relevance and coherence scores are strongly skewed toward 1 (positive), indicating that most steps within the solutions are contextually appropriate and logically consistent. The step-level correctness also has a high average but exhibits substantial variance, reflecting that some correct solutions still contain multiple locally incorrect steps. When the final answer is incorrect, relevance and coherence scores remain higher on average than correctness, suggesting that even a small fraction of locally irrelevant or incoherent steps can be critical enough to derail the overall solution. Moreover, they have notably higher variance than in correct-answer cases, indicating that while some solutions with wrong answers preserve logical flow until a late mistake, others collapse much earlier.

\begin{figure}[t]
    \centering
    \includegraphics[width=\linewidth]{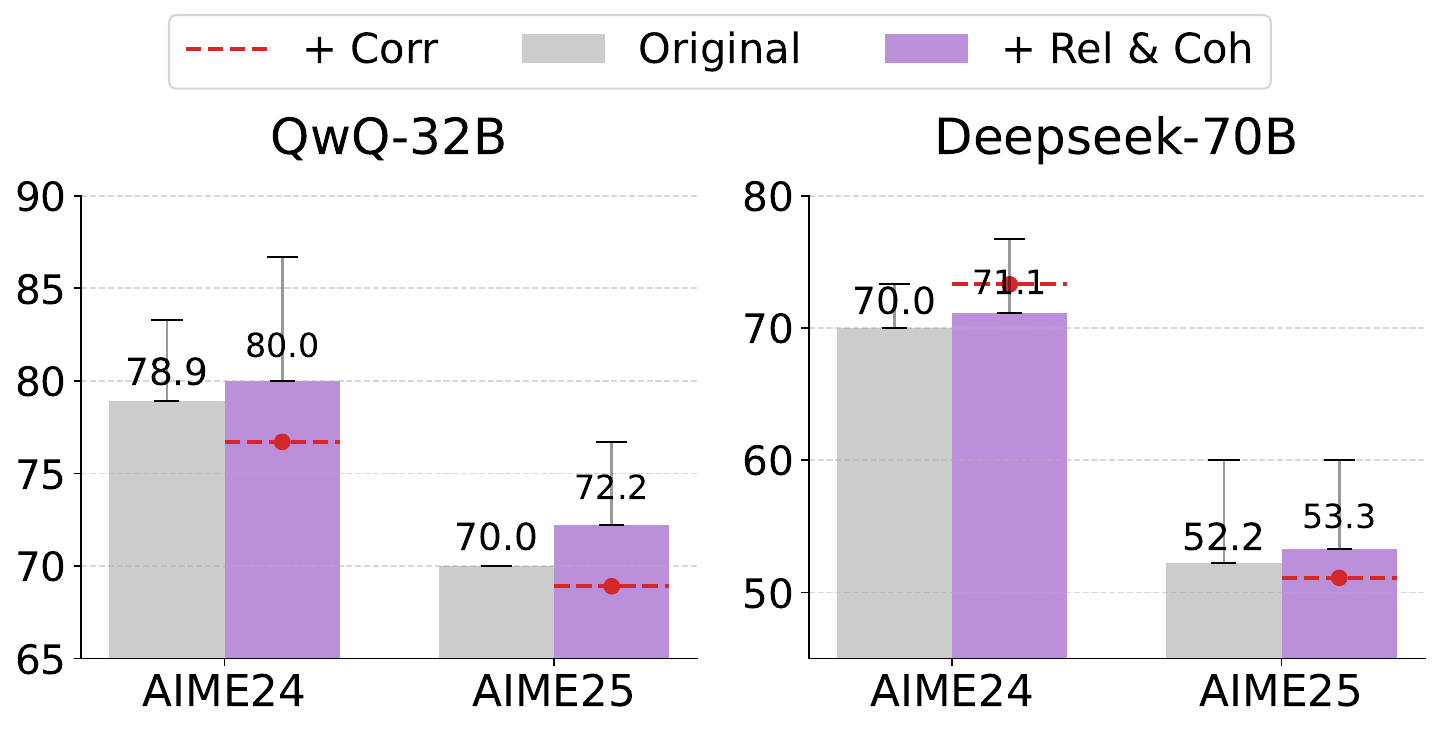}
    \caption{Average accuracy across three seeds on AIME24 and AIME25 for each method (left), and the prompt example used with QwQ-32B (right). The upper bound of each bar indicates the highest performance across the three runs, while the \textcolor[RGB]{209,15,15}{\dashuline{red dashed line}} shows the result when only correctness is emphasized (+Corr).}
    \label{fig:guide}
\end{figure}

\subsection{Aspect-Guided Inference} \label{sec4.2}
If relevance and coherence are meaningful indicators of reasoning quality, we hypothesized that emphasizing them during inference would enhance a model’s problem-solving ability. To validate this, we conduct an experiment with a lightweight prompting intervention that explicitly encourages LLMs to prioritize these aspects during step-by-step mathematical reasoning. We replace the baseline system prompt with an aspect-guided template that (i) defines \textit{relevance} and \textit{coherence} and (ii) instructs the model to satisfy both at every step, while keeping decoding settings and answer formatting unchanged (detailed prompts are in Appendix~\ref{appendix:guide}). As a control, we also test a \textit{correctness-guided} variant that emphasizes step-wise mathematical correctness only. Specifically, we apply this method to DeepSeek-R1-Distill-Llama-70B \cite{deepseekai2025deepseekr1incentivizingreasoningcapability} and QwQ-32B \cite{qwq32b}, both of which already achieve competitive performance on high school competition benchmarks, such as AIME24 and AIME25. 
For the baseline with the original condition, we adopt the one provided in the FuseAI \cite{wan2024knowledge} repository\footnote{\url{https://github.com/fanqiwan/FuseAI}} for Qwen-based and Deepseek-based models.
As shown in Figure~\ref{fig:guide}, even without any additional training, explicitly steering models toward relevance- and coherence-aware reasoning leads to noticeable gains in answer accuracy, +1.1 on average across models and datasets, echoing our earlier findings (§\ref{sec4.1}) that correct solutions tend to satisfy both criteria. These results imply that relevance and coherence can be actionable drivers of improved reasoning, further underscoring their potential as core dimensions that define reasoning quality.

\begin{figure}[t]
    \centering
    \includegraphics[width=\linewidth]{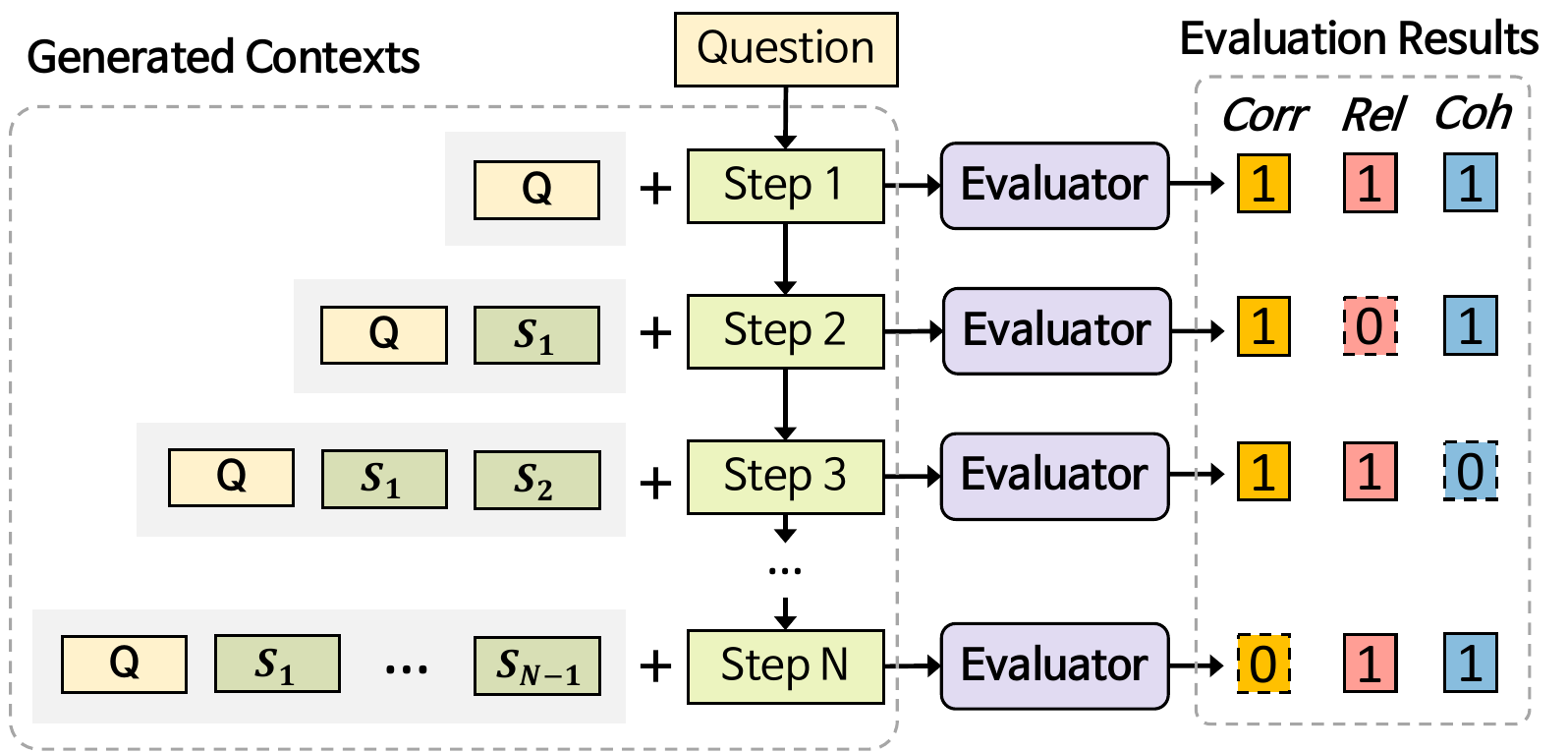}
    \caption{Overview of the CaSE method. LLM-based evaluators assess each reasoning step based on the preceding context, with respect to each aspect, such as \textit{Coherence} and \textit{Relevance}.
    }
    \label{fig:case}
\end{figure}

\begin{table*}[ht]
\centering
\footnotesize
\tabcolsep=0.4em
\resizebox{\textwidth}{!}{
\begin{tabular}{@{}llcccccc|cc|cccccc|cc@{}}
\toprule
\multirow{3}{*}{Model} & \multirow{3}{*}{Method} &
\multicolumn{8}{c|}{\textbf{MRa-GSM8K}} & 
\multicolumn{8}{c}{\textbf{MRa-MATH}} \\
\cmidrule(lr){3-10} \cmidrule(lr){11-18}
& & \multicolumn{2}{c}{\textbf{Relevance}} & \multicolumn{2}{c}{\textbf{Coherence}} & \multicolumn{2}{c|}{\textbf{Correctness}} & \multicolumn{2}{c}{\textbf{Average}}
& \multicolumn{2}{c}{\textbf{Relevance}} & \multicolumn{2}{c}{\textbf{Coherence}} & \multicolumn{2}{c|}{\textbf{Correctness}} & \multicolumn{2}{c}{\textbf{Average}}\\
& & Acc & F1 & Acc & F1 & Acc & F1 & Acc & F1 & Acc & F1 & Acc & F1 & Acc & F1 & Acc & F1\\
\midrule

\multirow{2}{*}{Phi-3.5-mini} 
& BoN  & {0.896} & 0.523 & 0.839 & 0.498 & 0.592 & 0.419 & 0.776 & 0.480 & 0.804 & 0.475 & 0.792 & 0.471 & 0.499 & 0.379  & 0.698 & 0.442 \\
& CaSE & 0.878 & {0.588} & {0.855} & {0.611} & {0.856} & {0.636} & \textbf{0.863} & \textbf{0.612} & {0.880} & {0.630} & {0.861} & {0.623} & {0.863} & {0.679} & \textbf{0.868} & \textbf{0.644}\\
\midrule
\multirow{2}{*}{Qwen2.5-7B} 
& BoN  & {0.805} & 0.599 & 0.844 & 0.627 & 0.632 & 0.594 &0.760 & 0.606 & {0.784} & {0.651} & 0.781 & 0.619 & 0.714 & 0.676 & 0.759 & 0.649\\
& CaSE & {0.805} & {0.626} & {0.846} & {0.666} & {0.897} & {0.687} & \textbf{0.849} & \textbf{0.660} & 0.767 & 0.628 & {0.795} & {0.642} & {0.884} & {0.736} & \textbf{0.815} & \textbf{0.669} \\
\midrule
\multirow{2}{*}{LLaMA3-8B} 
& BoN  & {0.887} & 0.581 & 0.871 & {0.610} & 0.632 & 0.530 &0.797 &0.574  & 0.814 & 0.552 & 0.814 & 0.574 & 0.645 & 0.552 & 0.758 & 0.559 \\
& CaSE & 0.863 & {0.651} & {0.886} & 0.592 & {0.874} & {0.607} & \textbf{0.874} & \textbf{0.617} & {0.861} & {0.618} & {0.889} & {0.629} & {0.868} & {0.657} & \textbf{0.873} & \textbf{0.635} \\
\midrule
\multirow{2}{*}{Qwen2.5-32B} 
& BoN  & 0.867 & 0.712 & {0.901} & {0.763} & 0.722 & 0.671 &0.830 &0.715  & 0.836 & 0.697 & {0.838} & {0.700} & 0.824 & 0.797 & 0.833 & 0.731 \\
& CaSE & {0.882} & {0.712} & 0.891 & 0.762 & {0.929} & {0.788} & \textbf{0.901} & \textbf{0.754} & {0.852} & {0.702} & 0.821 & 0.681 & {0.922} & {0.802} & \textbf{0.865} & \textbf{0.728}  \\
\midrule
\multirow{2}{*}{Qwen3-32B} 
& BoN  & 0.868 & 0.687 & 0.897 & 0.736 & 0.734 & 0.695 & 0.833 & 0.706 & 0.843 & {0.698} & 0.838 & 0.703 & 0.840 & {0.823} & 0.840 & 0.741  \\
& CaSE & {0.912} & {0.732} & {0.906} & {0.737} & {0.934} & {0.787} & \textbf{0.917} & \textbf{0.752} & {0.854} & 0.693 & {0.857} & {0.710} & {0.959} & {0.806} & \textbf{0.890} & \textbf{0.736}  \\
\midrule
\multirow{2}{*}{Qwen2.5-72B} 
& BoN  & 0.889 & 0.700 & 0.894 & 0.725 & 0.747 & 0.719 & 0.843 & 0.715 & {0.860} & 0.698 & 0.821 & 0.657 & 0.804 & 0.771 & 0.828 & 0.709  \\
& CaSE & {0.903} & {0.725} & {0.900} & {0.765} & {0.929} & {0.788} & \textbf{0.911} & \textbf{0.759} & 0.857 & {0.699} & {0.827} & {0.685} & {0.931} & {0.826} & \textbf{0.872} & \textbf{0.737} \\
\midrule
\multirow{2}{*}{GPT-4o} 
& BoN  & 0.897 & 0.733 & {0.912} & 0.759 & 0.903 & 0.745 & 0.904 & 0.746 & 0.870 & 0.704 & {0.838} & 0.695 & 0.821 & 0.798 & 0.843 & 0.732 \\
& CaSE & {0.915} & {0.737} & 0.903 & {0.772} & {0.927} & {0.788} & \textbf{0.915} & \textbf{0.766} & {0.873} & {0.711} & 0.836 & {0.696} & {0.922} & {0.820} & \textbf{0.877} & \textbf{0.742} \\

\bottomrule
\end{tabular}
}
\caption{Evaluation results on MRa-GSM8K and MRa-MATH datasets. This table presents the alignment between each model’s predictions and human judgments under two evaluation strategies (BoN and CaSE), measured by Accuracy and macro-F1 across three evaluation aspects.}
\label{tab:case_results}
\end{table*}

\section{Causal Stepwise Evaluation (CaSE)}
%Building on these empirical analyses, a natural next question is how to reliably evaluate relevance and coherence at the step level. 
Our analyses reveal the importance of relevance and coherence, motivating the need for a method that can automatically evaluate these aspects reliably at the step level. Accordingly, we propose Causal Stepwise Evaluation (CaSE), which assesses each reasoning step using only its preceding context (Figure~\ref{fig:case}). 
Unlike conventional paradigms, such as Best-of-N (BoN) or LLM-as-a-judge, which often expose the full reasoning trace, CaSE enforces a causal and incremental evaluation protocol on the evaluator model to prevent future information from influencing judgment. Although LLMs have recently shown promise in approximating human evaluators \cite{chiang2023can, wang2023chatgpt, fu2023gptscore}, most existing methods still rely on retrospective views, which can inflate perceived coherence or obscure early flaws. In contrast, CaSE restricts evaluation to only the context generated up to the current step, ensuring temporal grounding and causal consistency for judgment. Our design offers two key benefits: (1) It aligns with the stepwise generative process of LLMs, yielding evaluations that better reflect how models reason; (2) It avoids hindsight bias, enabling more accurate verification, supervision, and feedback. Overall, CaSE offers a principled framework for multi-aspectual step-level evaluation, establishing a solid foundation for analyzing and improving LLM reasoning through relevance and coherence.

\paragraph{Formulation}

Given a reasoning trace, i.e., solution steps before final answer, $\left[ \text{Step}_1, \dots, \text{Step}_N \right]$, which are generated in response to a question \( Q \), CaSE evaluates the \( k \)-th step with respect to an aspect \( a \in \{{Relevance}, {Coherence} \} \), by referring only to its preceding context and the given \( Q \):
\begin{equation}
\text{Eval}_{\text{aspect}}(\text{Step}_k \mid Q, C_{<k})
\end{equation}
where $C_{<k} = \{ \text{Step}_1, \dots, \text{Step}_{k-1} \}$ denotes the context prior to step $k$. We aim to ensure the evaluation reflects the local validity of each step, uninfluenced by future information or the final answer.

\paragraph{Experiments}
We test the effectiveness of CaSE as an automated reasoning evaluation framework on our proposed MRa benchmarks. Specifically, we examine whether instruction-tuned LLMs scaling from 3.5B to 72B parameters can reliably judge the quality of reasoning steps with respect to relevance and coherence under CaSE. As a baseline, we use the widely adopted BoN prompting strategy with $N=8$, where each reasoning trace is evaluated with access to the entire solution trace.

\section{Results}
\paragraph{Overall performance} Table~\ref{tab:case_results} presents step-level evaluation performance of CaSE and BoN across Relevance, Coherence, and Correctness, measured by the agreement between model predictions and human annotations in terms of Accuracy and macro-F1. Overall, CaSE consistently outperforms BoN across most models and aspects, with higher average scores observed on both MRa-GSM8K and MRa-MATH. This consistency across benchmarks highlights the generalizability of the CaSE evaluation framework. Notably, the performance gap between CaSE and BoN is more pronounced for smaller models such as Phi-3.5-mini and Qwen2.5-7B, which are more susceptible to information leakage from future steps when exposed to full reasoning traces. The substantial gains observed in these models suggest that enforcing a causal, context-restricted evaluation can effectively mitigate this issue. In contrast, stronger models like GPT-4o and Qwen2.5-72B exhibit smaller gaps between the two evaluation strategies, though CaSE still yields marginal improvements, underscoring its capacity to capture finer distinctions in reasoning quality even in high-performing models.

\begin{figure}[t]
    \centering
    \includegraphics[width=\linewidth]{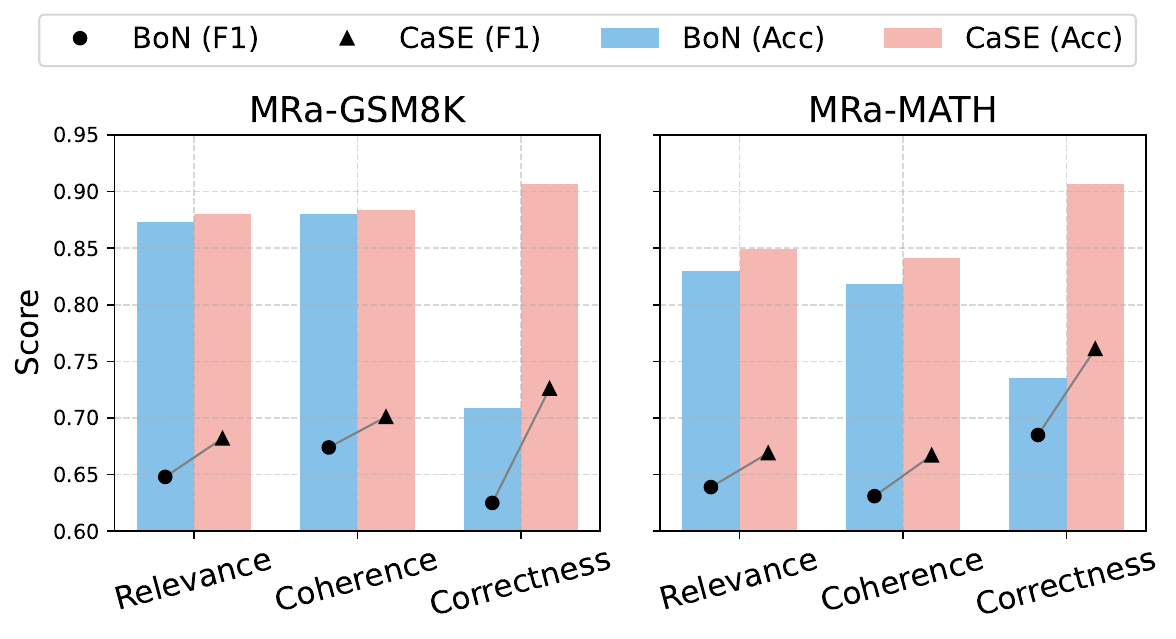}
    \caption{Evaluation results averaged across models for three evaluation aspects on MRa-GSM8K and MRa-MATH. Bars indicate Accuracy, and black dots represent macro-F1 under two evaluation strategies.
    }
    \label{fig:aspects_eval}
\end{figure}

\paragraph{Aspect-wise and model-wise performance}
Figure~\ref{fig:aspects_eval} summarizes model-averaged evaluation results across the three aspects. CaSE consistently outperforms BoN across all aspects and datasets, with the most substantial gains observed in Correctness, indicating that causal, step-specific evaluation better detects reasoning failures that may be overlooked when evaluating full-solution traces.

\begin{figure}
    \centering
    \includegraphics[width=\linewidth]{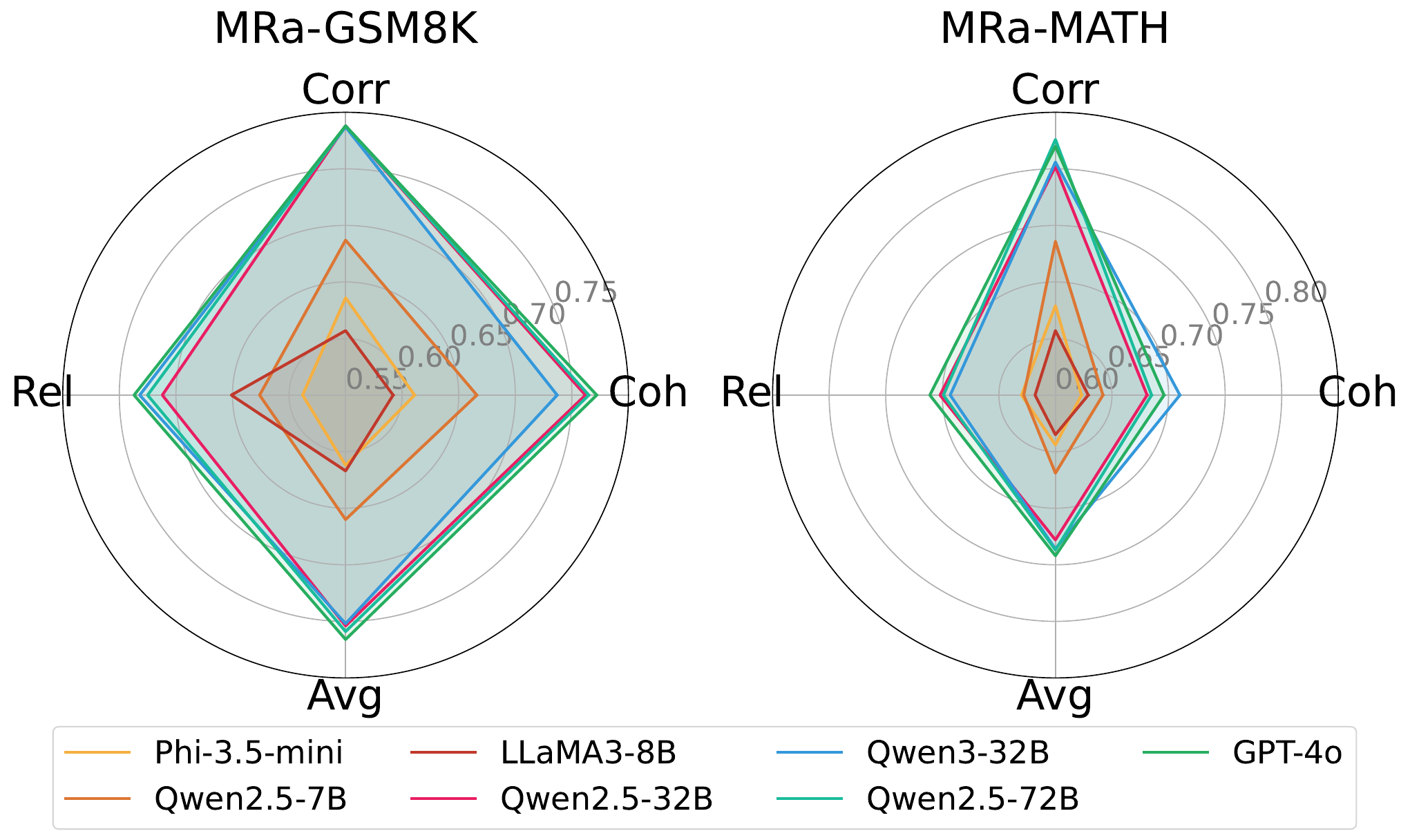}
    \caption{Model-wise F1-macro performance of CaSE across Relevance (Rel), Coherence (Coh), Correctness (Corr), and their Average (Avg) on two benchmarks.
    }
    \label{fig:model_wise}
\end{figure}

Figure~\ref{fig:model_wise} further breaks down the macro-F1 performance of CaSE by model. Larger and more capable models form wider polygons, indicating stronger alignment with human judgments across aspects. While CaSE yields reliable evaluations overall with the robust LLMs, we observe that Coherence and Relevance scores are generally lower than Correctness scores on MRa-MATH, even under CaSE. This result suggests that further refinement and discussions are necessary to reliably assess such nuanced reasoning dimensions, especially for more complex problem-solving scenarios.

\section{Discussions}

We leverage CaSE to investigate the practical value of evaluating reasoning quality via relevance and coherence. This includes curating supervised fine-tuning (SFT) data based on CaSE-evaluated multi-aspect scores and using it as an analytic tool to distangle the quality of model-generated reasoning traces.

\subsection{CaSE for SFT Data Curation}

Prior work shows that carefully curating small but high-quality datasets can substantially improve SFT performance~\cite{muennighoff2025s1, ye2025limo}. We explore whether CaSE can effectively support fine-tuning data curation by filtering individual reasoning steps or selecting entire samples that satisfy aspect-based quality criteria.

\begin{figure}[t]
    \centering
    \begin{subfigure}{\linewidth}
        \centering
        \includegraphics[width=1\linewidth]{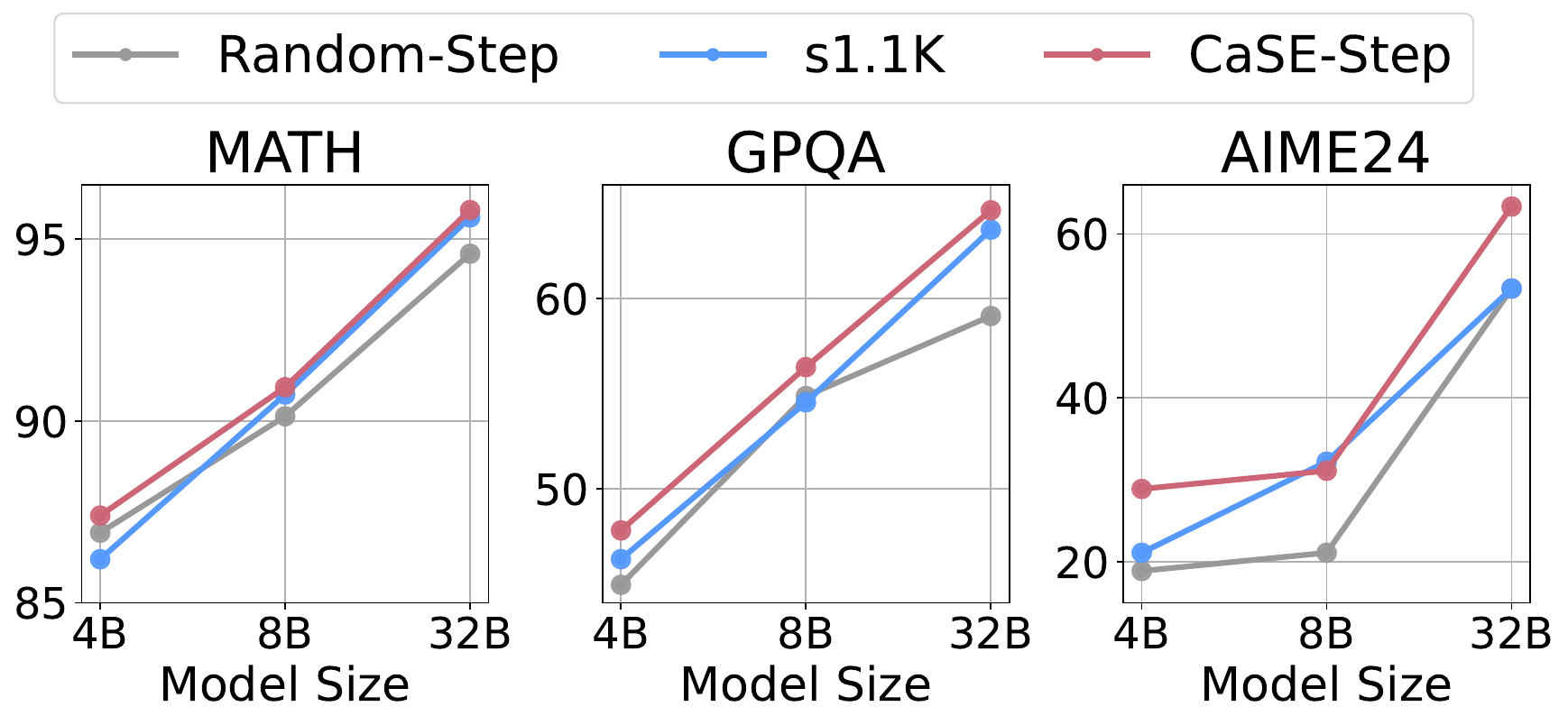}
        \caption{Step-level Filtering}
        \label{fig:sub1}
    \end{subfigure}
    
    \vspace{1em}
    
    \begin{subfigure}{\linewidth}
        \centering
        \includegraphics[width=1\linewidth]{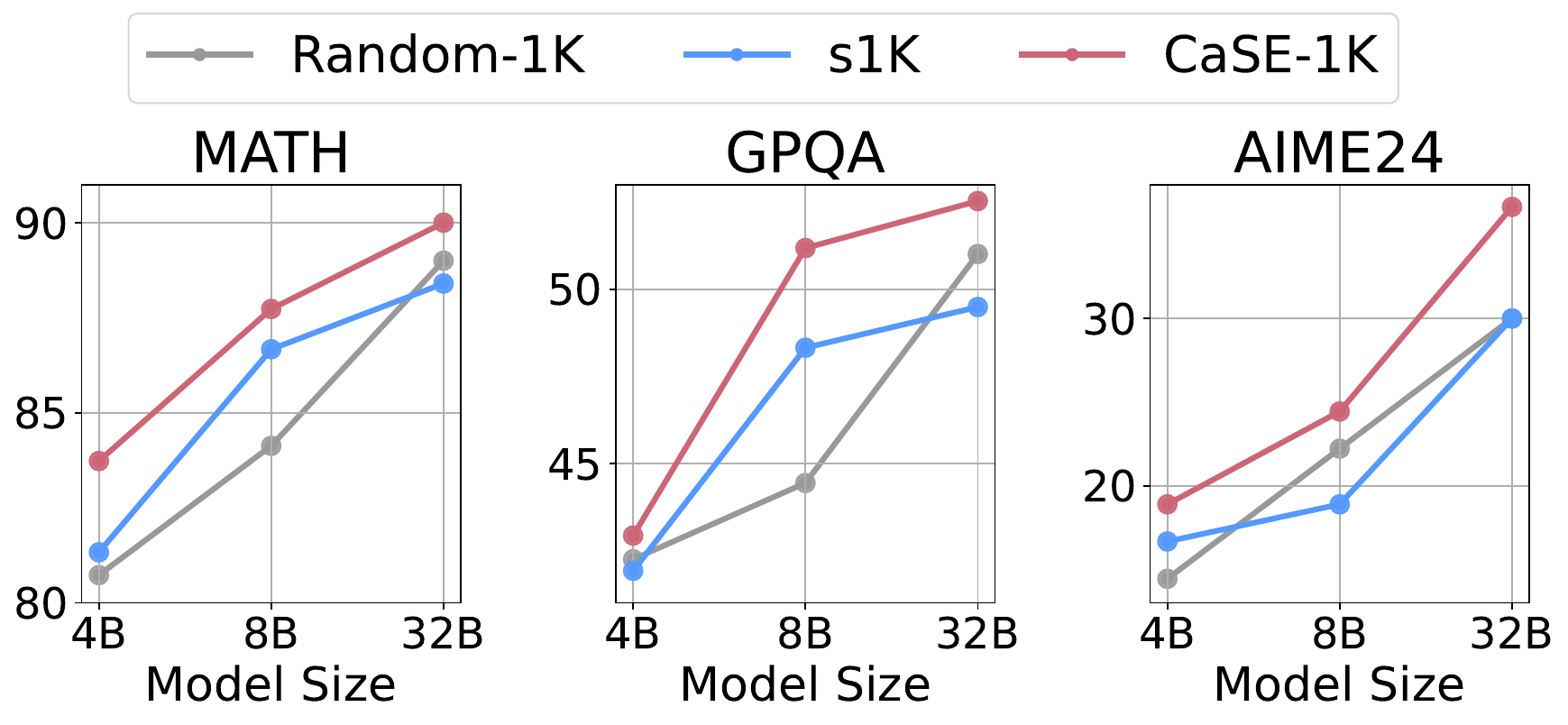}
        \caption{Sample-level Filtering}
        \label{fig:sub2}
    \end{subfigure}
    
    \caption{SFT performance: (a) with step-level CaSE filtering applied to s1.1K; (b) with sample-level CaSE filtering applied to the full Gemini-159K dataset, from which s1K was originally extracted.}
    \label{fig:s1_results}
\end{figure}

\paragraph{Experiments} Specifically, we compare CaSE-based filtering with two established baselines: s1K and s1K-1.1, which consist of 1K samples selected using human-defined heuristics such as difficulty, diversity, and overall quality~\citep{muennighoff2025s1}. The s1K dataset comprises thinking trajectories and solution steps generated by the Gemini 2.0 model, while s1K-1.1 is derived from DeepSeek-R1~\cite{deepseekai2025deepseekr1incentivizingreasoningcapability} outputs.

In our experiments, CaSE evaluates each step of the solution (or attempt) with respect to Relevance and Coherence, and curates data under two strategies (Figure~\ref{fig:s1_results}): (a) Step-level filtering, which prunes low-quality steps individually in the s1K-1.1 dataset; and (b) Sample-level filtering, which reconstruct 1K full solutions whose every step meets the quality criteria. We apply (a) directly to s1K-1.1, as it is more competitive than the s1K dataset. However, we apply (b) to the full 159K Gemini-generated data, from which s1K was originally curated, as it is the only available dataset with complete solution trajectories required for sample-level filtering. Note that while both approaches employ step-level evaluation, they differ in the granularity of filtering, \textit{i.e.}, one operates at the step level, the other at the solution level.

\paragraph{CaSE-based filtering yields consistent improvements}
Figure~\ref{fig:s1_results} demonstrates that CaSE-based filtering offers consistent advantages across all benchmarks (MATH, GPQA, AIME24) and model scales, outperforming both random selection and prior heuristic-based baselines, s1K and s1.1K. This highlights CaSE’s robustness and domain-general applicability as an automated and effective criterion for reasoning-focused data curation. Notably, even compared to s1.1K, which are carefully filtered through three-stage heuristics, CaSE achieves superior performance, establishing its strength as a scalable alternative to manual filtering. Among the two filtering strategies, sample-level filtering yields particularly large performance gains, indicating that retaining only fully coherent and relevant solution trajectories leads to higher alignment with reasoning quality. These gains are especially prominent in challenging benchmarks like AIME24 with smaller models (e.g., 4B), where fine-grained pruning is critical under limited supervision. Furthermore, larger models (e.g., 32B) appear better equipped to leverage the nuanced quality signals captured by CaSE, consistently achieving the highest performance across all benchmarks. Overall, these findings underscore CaSE’s promise as a principled and scalable tool for enhancing SFT data quality in complex reasoning tasks.

\begin{table}[t]
\centering
\scalebox{0.79}{
\begin{tabular}{l|ccc|ccc}
\toprule
 & \multicolumn{3}{c|}{\textbf{AIME24}} & \multicolumn{3}{c}{\textbf{AIME25}} \\
\textbf{Method} & \textbf{Corr} & \textbf{Coh} & \textbf{Rel} & \textbf{Corr} & \textbf{Coh} & \textbf{Rel} \\
\midrule
QwQ-32B        & 83.3 & 73.3 & 83.3 & 70.0 & \textbf{90.0} & \textbf{90.0} \\
+ MA Guide         & \textbf{86.7} & \textbf{83.3} & \textbf{90.0} & \textbf{76.7} & 86.7 & 86.7 \\
\midrule
Deepseek-70B   & 73.3 & 60.0 & 60.0 & 60.0 & 66.7 & 73.3 \\
+ MA Guide         & \textbf{76.7} & \textbf{66.7} & \textbf{76.7} & \textbf{60.0} & \textbf{76.7} & \textbf{86.7} \\
\bottomrule
\end{tabular}}
\caption{Evaluation of reasoning quality comparing the original inference with the one guided to focus on relevance and coherence (Multi-aspect Guide; \textit{MA Guide}), assessed via CaSE. Relevance and coherence scores are aggregated at the solution level. 
}\label{tab:tab2}
\end{table}

\subsection{Dissecting Reasoning Quality with CaSE}
\paragraph{Multi-aspect-guided inference results} 
In Section~\ref{sec4.2}, we observed that inference-time guidance on the proposed aspects improves problem-solving accuracy for reasoning-oriented models such as QwQ-32B and Deepseek-R1-70B. Then, do these gains reflect actual improvements in reasoning quality, specifically in terms of relevance and coherence? To answer this, we leverage CaSE as an evaluation metric to dissect how inference-time guidance shapes the quality of generated reasoning traces beyond correctness. For efficient inference, we use Qwen-32B-Instruct as the evaluator. As shown in Table~\ref{tab:tab2}, prompting models to prioritize relevance and coherence leads to improvements in both aspects on AIME24 and AIME25, except for QwQ-32B, which already achieved the highest performance of 90. These findings suggest that the accuracy gains are not superficial but stem from underlying improvements in reasoning quality. 

\paragraph{SFT results with CaSE-curated data} 
Complementing the correctness gains reported in Figure~\ref{fig:s1_results}, Table~\ref{tab:case_comparison} presents a fine-grained analysis of reasoning quality across relevance and coherence. Training on Case-1K filtered data not only enhances final answer accuracy but generally improves reasoning quality across model scales. For instance, the Qwen-2.5-32B-Instruct model trained on CaSE-1K outperforms its s1K counterpart by +6.66 in coherence and +13.33 in relevance, highlighting that CaSE-curated data fosters more logically consistent and contextually grounded reasoning. Notably, even smaller models such as 4B and 8B benefit from being trained with improved intermediate traces, suggesting that CaSE filtering effectively injects desirable inductive biases regardless of scale. These findings support the value of CaSE as a practical criterion for data selection to enhance task performance.

\begin{table}[t]
\centering
\scalebox{0.79}{
\begin{tabular}{llccc}
\toprule
\textbf{Model} & \textbf{Dataset} & \textbf{Corr} & \textbf{Coh} & \textbf{Rel} \\
\midrule
\multirow{2}{*}{4B}  & s1K       & 16.67 & 30.00 & 36.67 \\
                     & CaSE-1K   & \textbf{18.89} & \textbf{36.67} & \textbf{42.22} \\
\midrule
\multirow{2}{*}{8B}  & s1K       & 18.89 & 40.00 & \textbf{47.78} \\
                     & CaSE-1K   & \textbf{24.44} & \textbf{41.11} & 45.56 \\
\midrule
\multirow{2}{*}{32B} & s1K       & 30.00 & 46.67 & 46.67 \\
                     & CaSE-1K   & \textbf{36.67} & \textbf{53.33} & \textbf{60.00} \\
\bottomrule
\end{tabular}
}
\caption{Complementary to the accuracy results of sample-level in Figure 9 (b), this table reports reasoning quality along two additional dimensions, relevance and coherence, aggregated at the solution level.
}
\label{tab:case_comparison}
\end{table}

\section{Conclusion}
We introduced a stepwise, multi-aspect framework for evaluating LLM reasoning beyond correctness, focusing on relevance and coherence. Analyses on the proposed MRa-GSM8K and MRa-MATH show that these aspects provide complementary insights and, when emphasized at inference time, improve accuracy. To enable automated evaluation, we presented CaSE, a causal step-level method that better aligns with human judgments; further, CaSE-based SFT data curation notably improves LLM performance on math benchmarks. Overall, our findings establish multi-aspect stepwise evaluation as a practical foundation for advancing LLM reasoning.

\bibliography{tacl2021, references}
\bibliographystyle{acl_natbib}

\vspace{85mm}

\appendix

\section{Aspect-guided Prompts}
\label{appendix:guide}

\begin{figure}[h]
    \centering
    \includegraphics[width=\linewidth]{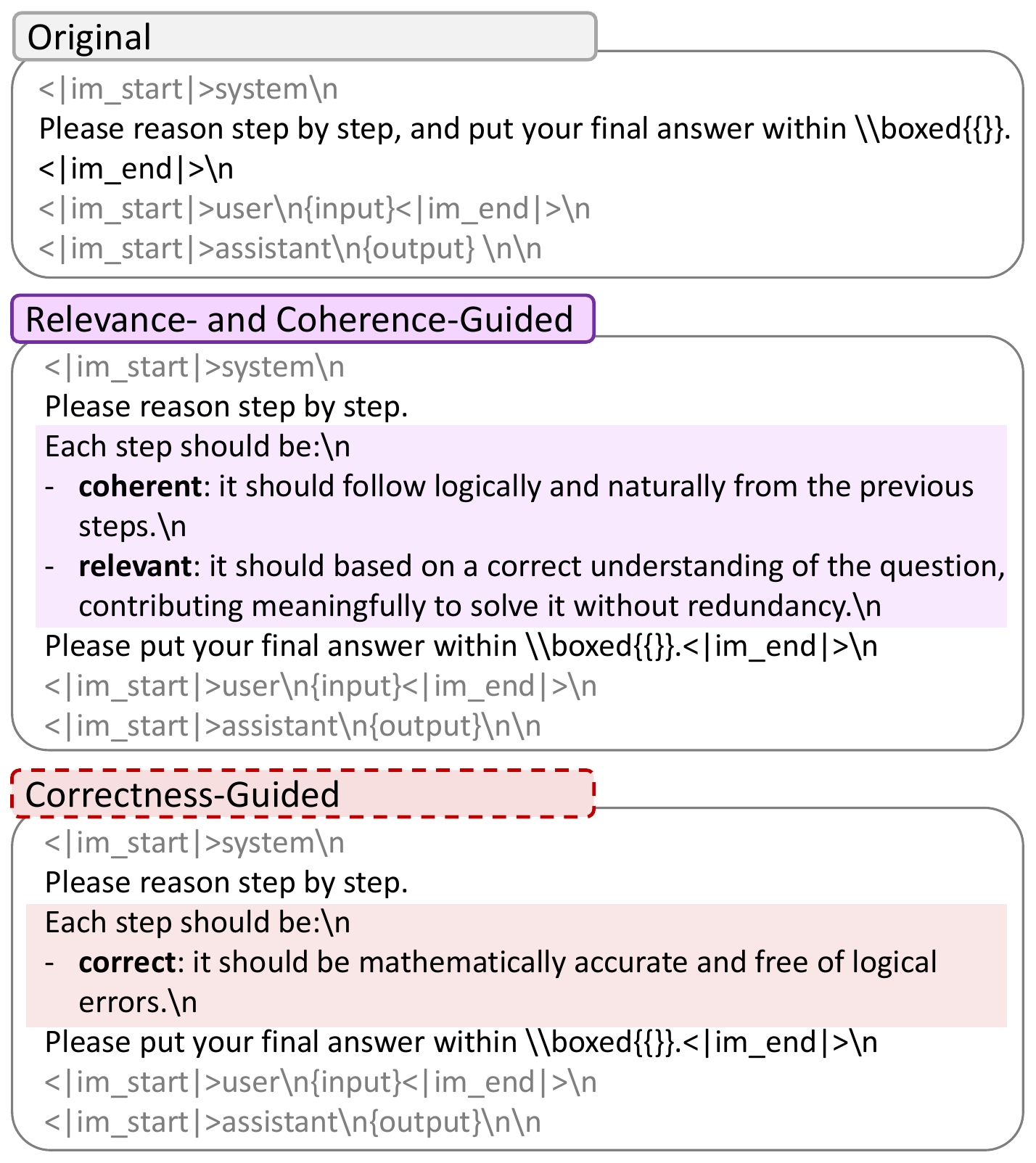}
    \caption{Prompts used for aspect-guided inference (Figure~\ref{fig:guide}). The example shown is for QwQ-32B; while details differ for DeepSeek-70B, the added (highlighted) phrases are identical and are based on the original prompts provided by FuseAI. 
    }
\label{fig:guide_prompts}
\end{figure}

\iftaclpubformat

\onecolumn
%\appendix

\fi

\end{document}